\documentclass[letterpaper, 10pt, conference]{IEEEtran}
\usepackage{times}

\usepackage[table]{xcolor}
\usepackage[numbers]{natbib}
\usepackage{multicol}
\usepackage[bookmarks=true]{hyperref}
\usepackage{graphicx}
\usepackage{amsmath}
\usepackage{amsfonts} 
\usepackage{booktabs} 
\usepackage{tabularx,array}
\usepackage{multicol}
\usepackage{multirow}
\usepackage{adjustbox}
\usepackage{tikz}
\usepackage{subcaption}
\usepackage{siunitx}
\usepackage{balance} 
\usepackage{bm}
\usepackage{makecell}

\usepackage{floatrow}
\definecolor{somegray}{rgb}{0.5, 0.5, 0.5}
\newcommand{\darkgrayed}[1]{\textcolor{somegray}{#1}}

\makeatletter
\newcommand*\titleheader[1]{\gdef\@titleheader{#1}}
\AtBeginDocument{%
  \let\st@red@title\@title
  \def\@title{%
    \vskip-2.0em
    \bgroup\normalfont\large\centering\@titleheader\par\egroup
    \vskip1em\st@red@title}
}

\makeatother

\titleheader{\darkgrayed{This abstract has been accepted for the Robotics: Science and Systems (RSS) Pioneers Workshop, 2024.}}

\title{Agile Robotics: Optimal Control,
Reinforcement Learning, and Differentiable Simulation}

\begin{document}


\author{Yunlong Song and Davide Scaramuzza \\
Robotics and Perception Group, University of Zurich}

\maketitle

\IEEEpeerreviewmaketitle
\begin{table*}[!h]
\centering
\setlength\tabcolsep{4pt} 
\renewcommand*{\arraystretch}{1.2}
\fontsize{7pt}{8pt}\selectfont
\begin{adjustbox}{max width=\textwidth}
\begin{tabular}{c  c  c  c}
 & \multicolumn{3}{c}{\cellcolor{gray!80} \textbf{Continuous Time Optimal Control Problem}} \\
 & \multicolumn{3}{c}{ \makecell{Minimize a cost function over a time horizon: \\
 $ \min_{x(\cdot), u(\cdot)} \int_{0}^{T} \ell(x(t), u(t), t) \, dt + \ell(x(T)) $}} \\
 \hline 
 \cellcolor{gray!20} \textbf{Control Method} 
 & \cellcolor{gray!20}\textbf{Model Predictive Control} 
 & \cellcolor{gray!20}\textbf{Policy Search}
 & \cellcolor{gray!20}\textbf{Backpropagation Through Time}\\
 \hline
 \textbf{Optimization Objective} 
 & \makecell{
   $ J(x, u) = \sum_{k=0}^{N-1} \ell(x_k, u_k) + \ell(x_N) $ \\
 }
  & \makecell{
   $ J(\theta) = \mathbb{E}_{\tau \sim \pi_{\theta}} \left[ \sum_{k=0}^{N} 
   r_k \right] $ \\
  }
  & \makecell{
   $ J(\theta) = \mathbb{E}_{x_0\sim p(x_0)} \left[ \sum_{k=0}^{N-1} 
   \ell(x_k, u_k) + \ell(x_N) \right] $ \\
  }
  \\
 \textbf{Constraints} 
  & \makecell{
    s.t. $ \begin{cases}
     x_0 = x_\text{init} \\
     x_{k+1} = f(x_k, u_k) \\ 
     g(x, u) = 0 \\ 
     h(x, u) \leq 0 
  \end{cases} 
   $
  }
  & \makecell{
    -
   }
  & \makecell{
    -
  }
  \\
  \textbf{Decision Variables}
  & \makecell{
    $ u_k, x_k $
  }
  & \makecell{
    $ \theta $
  }
  & \makecell{
    $ \theta $
  }
  \\
  \hline
  \textbf{Optimization Method}
  & \makecell{
    Nonlinear Programming
  }
  & \makecell{
    Policy Gradient
  }
  & \makecell{
    Analytical Gradient
  } 
  \\
  \textbf{Control Law}
  & \makecell{
    $ u_0^{\ast}$
  }
  & \makecell{
    $ u \sim \pi_{\theta^{\ast}}(u | x) $
  }
  & \makecell{
    $ u = \pi_{\theta^{\ast}}(x) $
  }
  \\
\bottomrule
\end{tabular}
\end{adjustbox}
\caption{\textbf{Comparison of three methods for approximately 
solving the continuous-time optimal control problem.} }
\label{tab:rl_oc_ds}
\end{table*}

Control systems are at the core of every real-world robot. They are deployed in an ever-increasing number of applications, ranging from autonomous racing and search-and-rescue missions to industrial inspections and space exploration.
To achieve peak performance, certain tasks require pushing the robot to its maximum agility. 
How can we design control algorithms that enhance the agility of autonomous robots and maintain robustness against unforeseen disturbances? 
%
%
My research addresses this question by leveraging fundamental principles in optimal control, reinforcement learning, and differentiable simulation.

Optimal Control~\cite{arthur1975applied, bertsekas2012dynamic}, such as Model Predictive Control~(MPC), relies on using an accurate mathematical model within an optimization framework and solving complex optimization problems online.
Reinforcement Learning~(RL)~\cite{sutton2018reinforcement} optimizes a control policy to maximize a reward signal through trial and error.
Differentiable Simulation~\cite{suh2022differentiable, de2018end} promises better convergence and sample efficiency than RL by replacing zero-order gradient estimates of a stochastic objective with an estimate based on first-order gradients.
An overview of these three approaches is summarized in Table~\ref{tab:rl_oc_ds}. 

Particularly, model-free RL has recently achieved impressive results, demonstrating exceptional performance in various domains, such as
autonomous drone racing~\cite{song2021autonomous, song2023reaching, kaufmann2023champion} and quadrupedal locomotion over challenging terrain~\cite{hwangbo2019learning,lee2020learning,miki2022learning}.
Some of the most impressive achievements of RL are beyond the reach of existing optimal control (OC) systems. 
However, most studies focus on system design; less attention has been paid to the systematic study of fundamental factors that have led to the success of RL or have limited OC. 

It is important to highlight that the progress in applying RL to robot control is primarily driven by the enhanced computational capabilities provided by GPUs rather than breakthroughs in the algorithms.
Consequently, researchers may resort to alternative strategies such as imitation learning~\cite{chen2020learning} to circumvent these limitations in scenarios where data collection cannot be accelerated through computational means~\cite{agarwal2023legged, fu2023deep, cheng2023parkour, zhuang2023robot}. 
This highlights the need to study the connection between RL, optimal control, and robot dynamics.
I attempt to answer the following three research questions: 

\textbf{Research Question 1}: \emph{What are the intrinsic benefits of reinforcement learning compared to optimal control?}

\textbf{Research Question 2}: \emph{How to combine the advantage of reinforcement learning and optimal control?}

\textbf{Research Question 3}: \emph{How to effectively leverage the dynamics of robots to improve policy training?}

\subsection{Reinforcement Learning versus Optimal Control}
\begin{figure}[htp]
     \centering
     \includegraphics[width=0.8\textwidth]{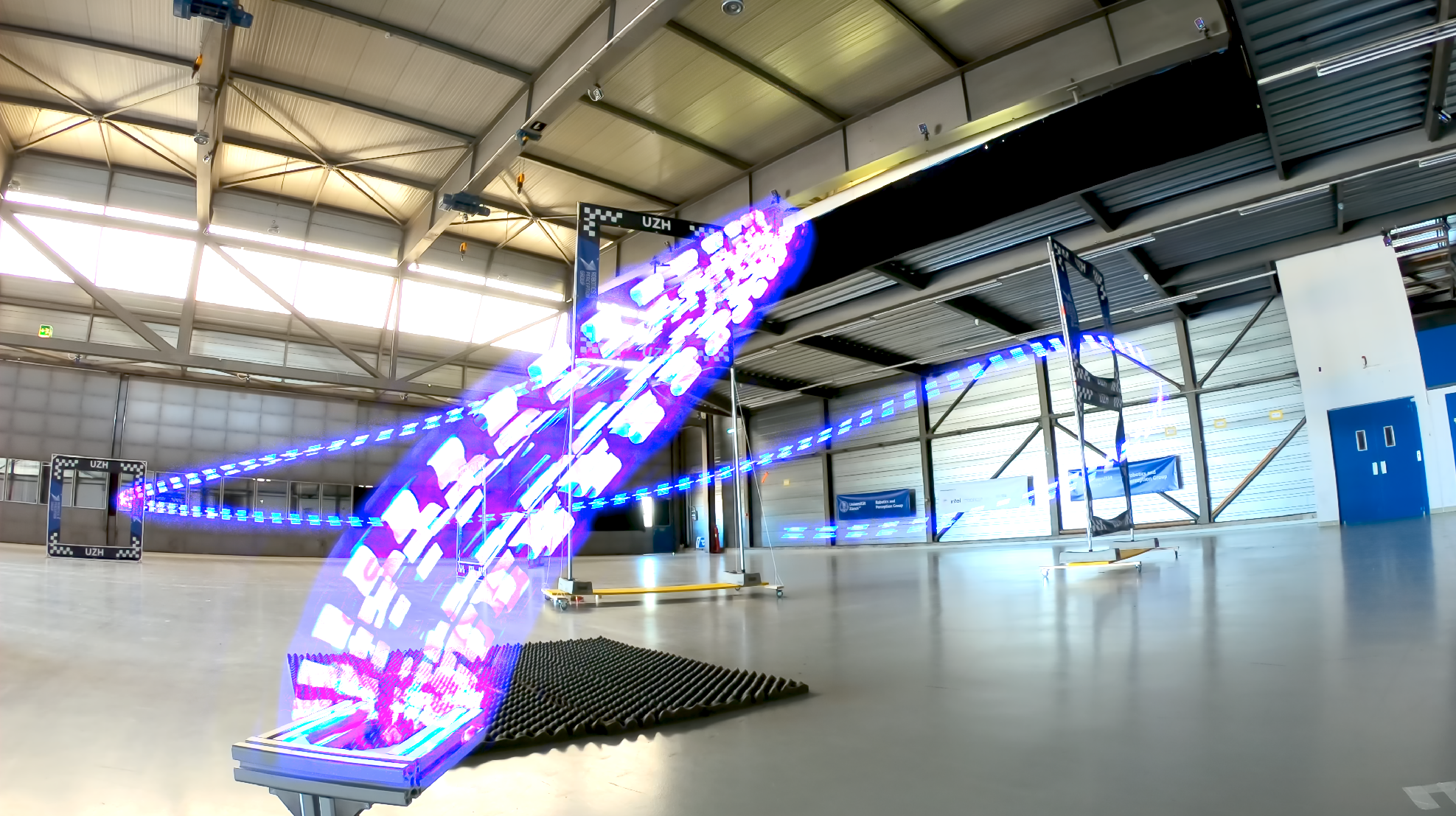}
     \caption{RL outperforms optimal control in drone racing~\cite{song2023reaching}.}
     \label{fig: racing}
\end{figure}
In~\cite{song2023reaching}, we investigate \textbf{Research Question 1} by studying RL and OC from the perspective of the optimization method and optimization objective. 
We perform the investigation in a challenging real-world problem that involves a high-performance robotic system: autonomous drone racing. 

On one hand, RL and OC are two different optimization methods and we can ask which method can achieve a more robust solution given the same cost function.
On the other hand, given that RL and OC address a given robot control problem by optimizing different objectives, we can ask which optimization objective can lead to more robust task performance. 

Our results indicate that the fundamental advantage of RL over OC lies in its optimization objective.
Specifically, RL directly maximizes a task-level objective, which leads to more robust control performance in the presence of unmodeled dynamics and disturbance.
In contrast, OC is limited by the requirement of optimizing a smooth and differentiable loss function, which in turn requires decomposing the task into planning and control, thus limiting the range of control policies that can be expressed by the system.
In addition, RL can leverage domain randomization to achieve extra robustness and avoid overfitting, where the agent is trained on a variety of simulated environments with varying settings. 

Our findings allow us to push an extremely agile drone to its maximum performance, achieving a peak acceleration greater than 12g and a peak velocity of 108~\SI{}{\kilo\meter\per\hour}. 
%
We show that the RL-based neural network policy outperforms state-of-the-art OC-based methods~\cite{Foehn2021science, Romero2021arxiv} in terms of robustness and lap time because RL does not rely on pre-computed trajectory or path. 
Fig~\ref{fig: racing} displays time-lapse illustrations of the racing drone controlled by our RL policy in an indoor flying arena.

\subsection{Policy Search for Model Predictive Control}
\begin{figure}[htp]
     \centering
     \includegraphics[width=0.55\textwidth]{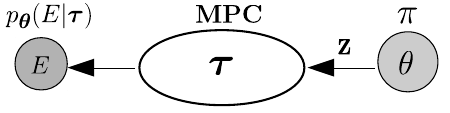}
         \caption{Graphical model of policy search for MPC.}
     \label{fig: vi}
\end{figure}

In~\cite{yunlong2022policy, song2020learning}, we investigate \textbf{Research Question 2} by presenting a \emph{policy-search-for-model-predictive-control} framework for merging
learning and control.  
A visualization of the framework is given in Fig~\ref{fig: vi}.
We consider model predictive control (MPC) as a parameterized controller and formulate the search for hard-to-optimize decision variables as a probabilistic policy search problem. 
Given the predicted decision variables, MPC solves an optimization problem and generates control commands for the robot.
A key advantage of our approach over the standard MPC formulation is that the high-level decision variables, which are difficult to optimize simultaneously with other state variables, can be learned offline and selected adaptively at runtime.

We validate this framework by focusing on a challenging problem in agile drone flight: flying a quadrotor through fast-moving gates. 
Flying through fast-moving gates is a proxy task to develop autonomous systems that can navigate the vehicle through rapidly changing environments. 
Our controller achieved robust and real-time control performance in both simulation and the real world. 
Additionally, this framework can be used for controller tuning~\cite{romero2023weighted}. 

\subsection{Policy Learning via Differentiable Simulation}
In~\cite{song2024learning}, we investigate \textbf{Research Question 3} by demonstrating the effectiveness of differentiable simulation for policy training. Differentiable simulation promises faster convergence and more stable training by computing low-variant first-order gradients using the robot model, but so far, its use for robot control has remained limited to simulation~\cite{xu2021accelerated, ren2023diffmimic, huang2021plasticinelab, freeman2021brax}.

\begin{figure}[htp]
\centering
\includegraphics[width=0.7\textwidth]{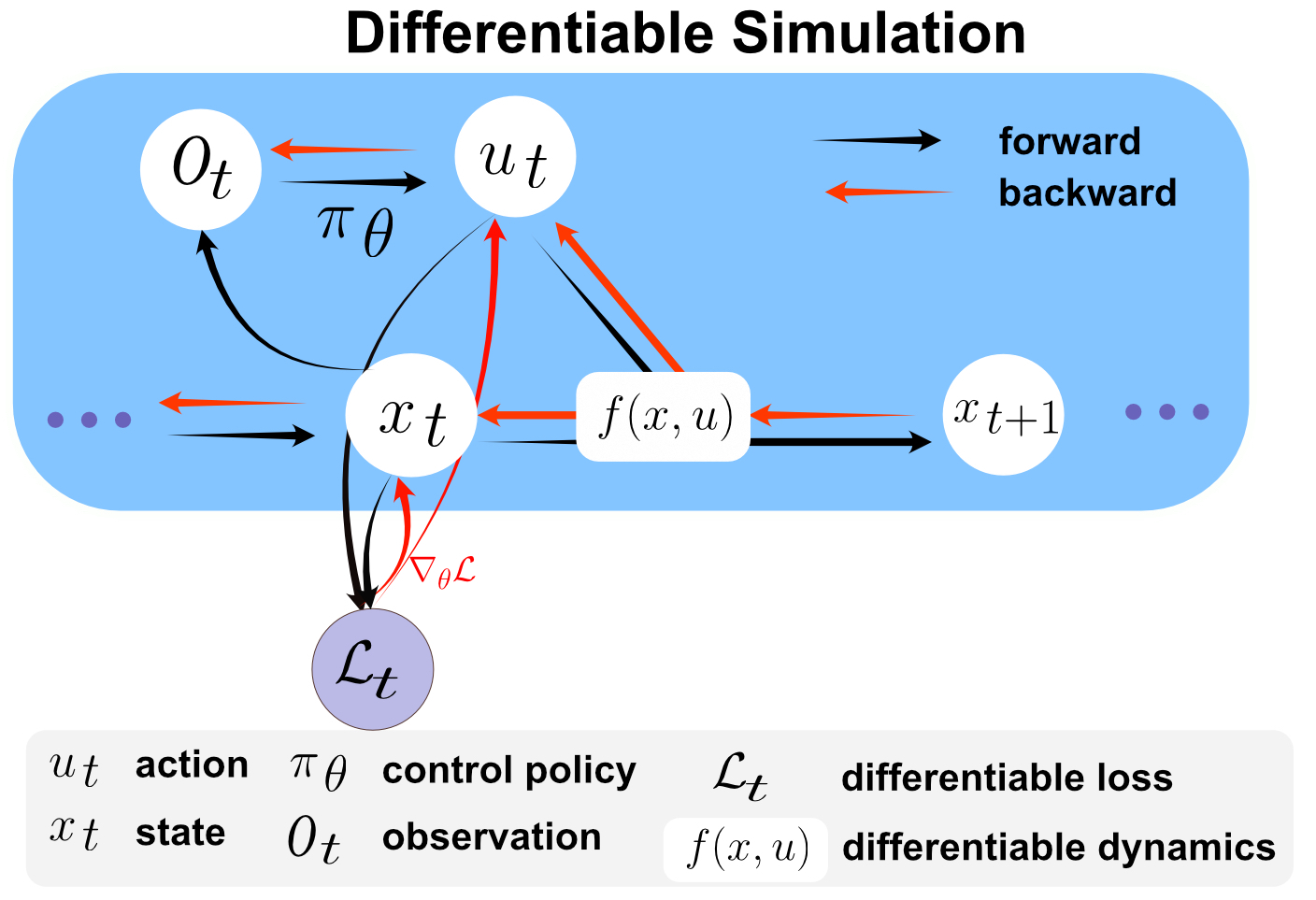}
\caption{Graphical model of Differentiable Simulation. }
\label{fig:diff_sim}
\end{figure}
\vspace{-1pt}

In~\cite{song2024learning}, we tackle the challenge of learning control policies for quadruped locomotion. 
The main challenge with differentiable simulation lies in the complex optimization landscape of robotic tasks due to discontinuities in contact-rich environments, particularly quadruped locomotion. 
We propose a new, differentiable simulation framework to overcome these challenges. 
The key idea involves decoupling the complex whole-body simulation, which may exhibit discontinuities due to contact, into two separate continuous domains. 
Our framework enables learning quadruped walking in minutes using a single simulated robot without any parallelization.
When augmented with GPU parallelization, our approach allows the quadruped robot to master diverse locomotion skills, including trot, pace, bound, and gallop, on challenging terrains in minutes.
Additionally, our policy achieves robust locomotion performance in the real world zero-shot.
To the best of our knowledge, this work represents the first demonstration of using differentiable simulation for controlling a real quadruped robot.
This work provides several important insights into using differentiable simulations for legged locomotion in the real world.

\subsection{Future Work}
Incorporating structured knowledge from robot dynamics and constraints from optimal control into reinforcement learning could potentially reduce the sample complexity and improve learning efficiency. 
This could involve using optimal control as parameterized policy~\cite{yunlong2022policy} or as an implicit differentiable layer~\cite{amos2018differentiable, romero2023actor} or integrating physical laws and safety constraints directly into the learning process. 

In my future research, I plan to focus on developing advanced control frameworks that merge the precision and safety of optimal control with the adaptability and robustness of reinforcement learning.
%
%
I plan to focus on more challenging robot control tasks, including vision-based humanoid locomotion. 
My ultimate objective is to achieve a level of locomotion performance comparable to that of a human, navigating through challenging terrains while avoiding obstacles.

\newpage

\balance
\bibliographystyle{plainnat}
\bibliography{references}

\end{document}